# Evaluation of YOLO Models with Sliced Inference for Small Object Detection

Muhammed Can Keleş[1], Batuhan Salmanoğlu[1], Mehmet Serdar Güzel, Baran Gürsoy[1], Gazi Erkan Bostancı[1]

[1]Department of Computer Engineering, Ankara University, Ankara, Turkey

*Abstract*—Small object detection has major applications in the fields of UAVs, surveillance, farming and many others. In this work we investigate the performance of state of the art Yolo based object detection models for the task of small object detection as they are one of the most popular and easy to use object detection models. We evaluated YOLOv5 and YOLOX models in this study. We also investigate the effects of slicing aided inference and fine-tuning the model for slicing aided inference. We used the VisDrone2019Det dataset for training and evaluating our models. This dataset is challenging in the sense that most objects are relatively small compared to the image sizes. This work aims to benchmark the YOLOv5 and YOLOX models for small object detection. We have seen that sliced inference increases the AP50 score in all experiments, this effect was greater for the YOLOv5 models compared to the YOLOX models. The effects of sliced fine-tuning and sliced inference combined produced substantial improvement for all models. The highest AP50 score was achieved by the YOLOv5-Large model on the VisDrone2019Det test-dev subset with the score being 48.8.

## i. Introduction

Yolo based models have been used widely in object detection problems due to their ease of use, ease of training and high performance. Many state of the art object detection models are tested on benchmark datasets such as COCO [1] and Pascal [2], while these datasets are of high quality, they fail to represent the problems that would be faced in many applications of object detection to real world problems. We can list aerial object detection, surveillance applications and farming applications as examples where detection of small objects would be crucial. We use a slicing aided inference method proposed at [3] with the aim to improve the performance of the models. SAHI is an slicing aided inference method for object detection and instance segmentation models that performs inference by processing the images by crops. We have also tested training the models with a modified version of the VisDrone2019Det [4] dataset where we crop the images to smaller sizes to improve the performance gain that would come from using SAHI. We evaluated the YOLOv5 [5] model, which is an improved version of the YOLOv4 [6] model, and the YOLOX [7] model, which is an extension of the YOLO model series and is an anchor free model. We tested two different models and two different versions of those models, totaling four models. We test the YOLOv5-Small, YOLOv5-Large, YOLOX-Small, YOLOX-Large models, each for four different experimental settings.

## ii. Related work

YOLO based models are widely used because of the reasons stated above. There were some works [8] that aimed to improve the performance of the YOLO models for small object detection. While there are many models that achieve better scores for small object detection than YOLOv5 and YOLOX models, the ease of using and ease of training these models makes them more desirable to work with for general applications.

The authors at [3] have experimented with VFNet [9], TOOD [10] and FCOS [11] models and they have reported the effects of sliced fine tuning and slicing aided hyper inference for their models. Authors at [3] have shown that slicing aided hyper inference increases the small object detection performance, though in some cases the performance for the large object decreases. They also have shown that sliced fine tuning increases the performance for small object detection. It should be noted that sliced inference increases the inference time of the models because of the increased amount of data that has to be processed by the models. It should also be noted that the authors at [3] have used a number of extra steps to improve the AP score, such as combining SAHI and full inference.

TOOD aims to learn an intersection-over-union-aware classifier to overcome the issue of ranking predictions in object detection problems. This problem is crucial in dense object detection settings. The authors state that an AP score of %51,1 was achieved on the COCO dataset. The authors at [3] have reported an AP50 score of %43.5 for the highest scoring experiment on the VisDroneDet-2019 test-dev subset.

FCOS is an anchor-box-free object detection model that relies on per-pixel detection similar to semantic segmentation. The authors of the paper state that the model archives an AP score of %44,7 on the COCO dataset. FCOS has been shown to achieve %38,5 AP over the VisDrone2019Det test-dev set by the authors of the SAHI paper.

VFNet is a model that builds on the FCOS model. The VTNet model predicts a star shape bounding box in order to perform better at dense object detection settings. The authors at [9] state that an AP score of %55,1 was achieved on the COCO dataset. VFNet has been shown to achieve an AP50 score of %42,2 on the VisDroneDet-2019 test-dev sets by the authors at [3].

YOLOv5 is an extension to the YOLOv4 model that is highly used in a variety of object detection applications. It is a single stage object detection model that is pre-trained on the COCO dataset and different versions of the models are accessible here [5]. The model uses a CSPNet [12] as a backbone and a PANet [13] is used as a neck to get feature pyramids. The YoloV5-Large model has been shown to achieve an AP score of %49,0 on the COCO dataset.

YOLOX is a version of the YOLO models that is anchor free and a version that uses some advanced techniques such as head-decoupling and leading label strategy. The paper reports higher AP scores on the COCO dataset compared to the YOLOv5 models. As the YOLOv5 model is improved continuously, this superiority doesn't seem to hold at this time. The authors of the YOLOX model reported an AP score of %50,0 on the COCO dataset.

YOLOF [14] proposes a single level feature map instead of a full FPN [15] layer as the feature extractor. The paper also points out that the success of the FPN's are due to divide and conquer strategy rather than the fusion of multi-scale feature maps that are produced by the FPN's. The authors have shown that the model archives an AP score of %44,3 on the COCO dataset.

In the VisDrone2021-Det [16] challenge results, it is reported that an AP50 score of 65.34 was achieved by the DBNet [16] model, the following best models were DBNet, VistrongerDet [17], SOLOer [16], Swin-T [18], EfficientDet [19], cascade++ [16], DNEFS [16], TPH-YOLOv5 [20] with AP50 scores of 65.34, 64.28, 63.91, 63.91, 63.25, 62.92, 62.86, 62.83. The DBNet model is based on the Cascade R-CNN [21] model. DBNet also utilizes Deformable Convolution [22] for the bottleneck. VistrongerDet is a two stage-detector that utilizes FPN's and the paper proposes a method that can be used with any other FPN based two stage detectors. SOLOer is a model that is based on YOLOv4 and it utilizes BiFPN's [19]. Swin-T is a vision transformer based object detector which uses a sliding-window approach to process images, the model archives an AP score of 58.7 on the COCO dataset. EfficientDet is the first work that introduces BiFPN's, the model aims to improve both accuracy and inference time for the object detection task. EfficientDet scores 55.5 AP points on the COCO dataset. Cascade++ is a model that is also based on the Cascade-RCNN model. It is an anchor-free model and it utilizes a sequence of detectors which use different IoU thresholds. DNEFS model is based on FPN, Cascade-RCNN and YOLOv5 models. The model also utilizes the SWA [23] training strategy to further improve the performance. TPH-YOLOv5 model is based on the YOLOv5 model and it utilizes both Convolutional Block Attention Module [24] for finding regions that would have dense objects.

### iii. Dataset

The VisDrone2019Det dataset was used for training and evaluating the models. The VisDrone dataset consists of 10,209 images, including 6,471 images in the training subset, 548 in the validation subset, 1,580 images in the test-challenge subset 1,610 images in the test-dev subset. The test-dev set was used for the evaluation of the models. There are ten classes in the dataset: *pedestrian, person, car, van, bus, truck, motor, bicycle, awning-tricycle, tricycle*. The distribution of the instances of each class can be seen in Figure 2. The pedestrian and person class possess the biggest challenge in regards to small object detection performance, due to the instances being small relative to the images and due to dense occurrences of these classes in crowded areas such as streets. An example for a crowded example from the VisDrone dataset can be seen at Figure 1.

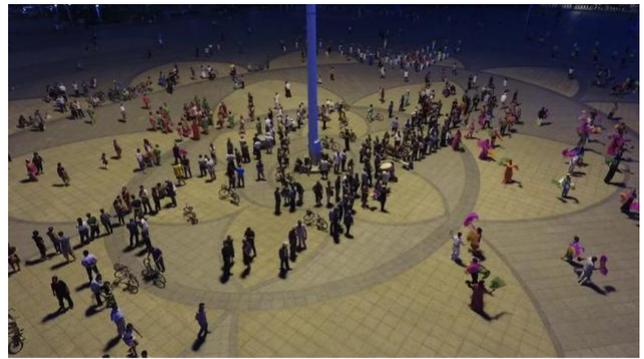

figure 1: An example image from the VisDroneDet-2019 dataset

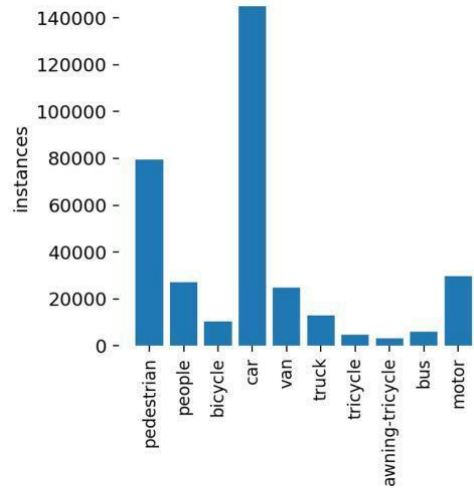

figure 2: Distribution of instances for the VisDrone dataset

The models for experiments 1 and 2 were trained on the original VisDrone images with a resolution of 1088 by 1088. The models for experiments 3 and 4 were trained on cropped parts of the original VisDrone images, the crops were 640 by 640 images with no overlap between the crops. The crops were resized to 1088 by 1088 for training the models.

When creating a cropped dataset, it is possible that some of the crops will have very big overlaps naturally even though we use 0 overlap between the crops. This is due to the shapes of the images. This would decrease the quality of the dataset because of the similarity between the near-duplicate examples. We can see an example of this in Figure 3. To avoid this problem, it is advised that you choose the width, height and overlap ratios of the crops carefully.

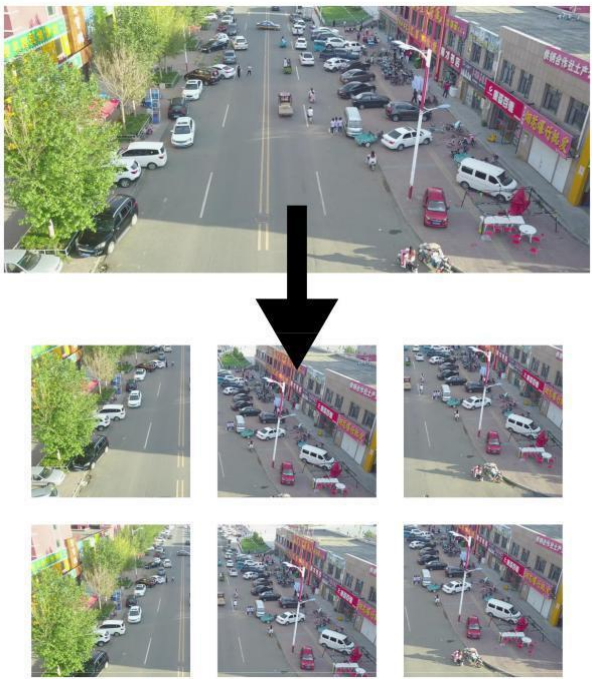

Figure 3: Example of instances of the cropped dataset

Another point that should be considered is some datasets contain classes such as "Ignored", this class usually represents far regions where the objects are really small and they do not have labels most of the time. When a cropped dataset is created, these regions will appear bigger but they won't have any labels in the training set. This could cause problems when there are many ignored regions in the dataset.

## IV. EXPERIMENTS AND METHODS

We trained and experimented with 2 different model types and 2 different model sizes for each model, totaling 4 models. We trained the YOLOv5-small, YOLOv5-large, YOLOX-small, YOLOX-large models in our work. The differences between the small and large variations of the models are the number of parameters. All of the models used in the experiments were pre-trained on the COCO Object detection dataset by the YOLOv5 and MMDetection [16] community. We used the Ultralytics YOLOv5 repository [5] to train the YOLOv5 models, and the MMDetection repository was used for training the YOLOX models. The slicing aided inference experiments were done using the SAHI repository [3].

Each model was trained on two different settings. In the first experimental setting the models were trained on 1088 by 1088 images from the VisDrone dataset, and in the second one the models were trained on a modified version of the VisDrone dataset where the images were cropped to 640 by 640 images. The cropped images were resized to 1088 by 1088 images while training the models. In total 8 models were trained.

For each model, the following 4 experiments were conducted.

*1- Training on full images and testing on full images*

This experiment aims to establish a baseline for evaluating the sliced inference and sliced fine tuning. The images were resized to 1088 by 1088 for training.

*2- Training on full images and SAHI inference*

This experiment aims to evaluate the effects of SAHI with no pre-training on the specific crop sizes the model would be evaluated on. The models used for this experiment were the same models as the ones in the first experiment.

*3- Training on cropped images and testing on full images*

This experiment aims to evaluate if the performance increase that is obtained by sliced fine tuning is coming from SAHI or does the model perform better in the cases where SAHI wasn't used.

*4- Training on cropped images and SAHI inference*

This experiment aims to evaluate the performance increase that the sliced inference and sliced fine tuning combined would give.

We trained the Yolov5 models with a learning rate of 0.01. The YoloX models were trained with a learning rate of 0.001. The YoloV5 models were trained using the ultralytics repository and the YoloX models were trained using the MMDetection repository.

The cropped dataset consisted of 640 by 640 crops from the VisDroneDet-2019 train-subset. The SAHI sliced predictions were done with 640 by 640 window size and 0.1 overlap between the windows. We used the AP50 score for evaluating all of the models.

Table 1: Classwise AP50 scores for experiments.

| Model | Dataset | Inference | Pedestrian | People | Bicycle | Car | Van | Truck | Tricycle | Awning-tricycle | Bus | Motor | All |
|---|---|---|---|---|---|---|---|---|---|---|---|---|---|
| YoloV5-Small | Standard | Standard | 34.2 | 18.3 | 12.3 | 76.0 | 29.2 | 40.9 | 14.4 | 11.6 | 52.9 | 33.4 | 32.0 |
| YoloV5-Small | Standard | SAHI | 44.4 | 22.6 | 16.2 | 79.0 | 37.2 | 37.7 | 18.6 | 13.9 | 54.9 | 37.0 | 36.2 |
| YoloV5-Small | Cropped | Standard | 26.3 | 16.3 | 7.9 | 69.6 | 25.0 | 35.5 | 13.2 | 6.8 | 49.4 | 26.2 | 27.6 |
| YoloV5-Small | Cropped | SAHI | 44.8 | 27.6 | 18.9 | 79.8 | 38.8 | 46.4 | 24.4 | 13.6 | 56.6 | 42.6 | 39.4 |
| YoloV5-Large | Standard | Standard | 42.5 | 23.6 | 18.9 | 80.2 | 36.8 | 50.8 | 24.7 | 17.3 | 62.4 | 43.1 | 40.0 |
| YoloV5-Large | Standard | SAHI | 51.4 | 30.5 | 22.7 | 82.2 | 43.3 | 49.9 | 29.9 | 19.6 | 64.0 | 47.3 | 44.1 |
| YoloV5-Large | Cropped | Standard | 38.2 | 22.9 | 15.4 | 76.6 | 37.3 | 47.9 | 21.7 | 22.4 | 60.6 | 36.8 | 38.0 |
| YoloV5-Large | Cropped | SAHI | **54.7** | **34.8** | **26.6** | **83.0** | 47.4 | **57.2** | **35.8** | **28.9** | **66.8** | **53.1** | **48.8** |
| YoloX-Small | Standard | Standard | 32.7 | 15.5 | 7.8 | 74.6 | 33.0 | 32.7 | 14.7 | 11.2 | 48.8 | 31.7 | 30.3 |
| YoloX-Small | Standard | SAHI | 38.4 | 21.1 | 9.5 | 74.5 | 35.4 | 30.5 | 15.6 | 11.6 | 45.6 | 30.0 | 31.2 |
| YoloX-Small | Cropped | Standard | 29.7 | 14.5 | 7.7 | 71.8 | 32.6 | 32.2 | 12.0 | 10.9 | 49.2 | 26.3 | 28.7 |
| YoloX-Small | Cropped | SAHI | 42.9 | 25.4 | 14.7 | 77.2 | 43.7 | 36.8 | 18.6 | 16.0 | 52.4 | 35.2 | 36.3 |
| YoloX-Large | Standard | Standard | 38.7 | 20.5 | 12.8 | 79.0 | 39.9 | 45.8 | 22.3 | 19.0 | 55.9 | 38.1 | 37.2 |
| YoloX-Large | Standard | SAHI | 42.7 | 25.7 | 14.9 | 78.5 | 43.7 | 42.8 | 22.6 | 18.7 | 51.6 | 32.9 | 37.3 |
| YoloX-Large | Cropped | Standard | 36.4 | 20.7 | 14.3 | 77.5 | 43.5 | 47.7 | 22.5 | 20.6 | 59.2 | 37.3 | 38.0 |
| YoloX-Large | Cropped | SAHI | 49.1 | 31.9 | 21.4 | 81.9 | **51.5** | 51.7 | 30.1 | 26.3 | 61.3 | 46.8 | 45.2 |

Table 2: AP50 scores for each bounding boxes size-wise.

| Model | Dataset | Inference | small | medium | large |
|---|---|---|---|---|---|
| YoloV5-Small | Standard | Standard | 20.7 | 47.1 | 50.5 |
| YoloV5-Small | Standard | SAHI | 26.4 | 47.5 | 52.3 |
| YoloV5-Small | Cropped | Standard | 14.9 | 44.5 | 61.1 |
| YoloV5-Small | Cropped | SAHI | 27.2 | 54.8 | 61.1 |
| YoloV5-Large | Standard | Standard | 27.2 | 56.2 | 66.4 |
| YoloV5-Large | Standard | SAHI | 33.2 | 57.5 | 63.9 |
| YoloV5-Large | Cropped | Standard | 23.8 | 56.3 | **71.4** |
| YoloV5-Large | Cropped | SAHI | **36.7** | **63.8** | 71.3 |
| YoloX-Small | Standard | Standard | 18.9 | 44.7 | 47.8 |
| YoloX-Small | Standard | SAHI | 21.8 | 43.6 | 45.4 |
| YoloX-Small | Cropped | Standard | 16.6 | 45.0 | 58.1 |
| YoloX-Small | Cropped | SAHI | 26.4 | 49.7 | 56.6 |
| YoloX-Large | Standard | Standard | 25.1 | 52.3 | 54.5 |
| YoloX-Large | Standard | SAHI | 26.4 | 51.2 | 53.5 |
| YoloX-Large | Cropped | Standard | 23.9 | 57.1 | 70.8 |
| YoloX-Large | Cropped | SAHI | 34.2 | 59.9 | 67.2 |

## V. RESULTS

We trained and tested the models as described in the experiments and methods section. The results at Table 1 show that large versions of both YOLOv5 and YOLOX models perform better than the small versions. We can also see that sliced inference improves the AP50 scores for both standard and cropped dataset training, while the models trained on the cropped dataset tend to benefit more from SAHI. The best performing model was the YoloV5-Large model, with an AP50 score of 48.8. This model was trained on the cropped dataset and was evaluated using sliced inference.

From Table 2, we see that the small object AP50 score of the YOLOv5-Large model trained on the cropped dataset and evaluated with SAHI is the highest, with a score of 36.7. We also see that the AP50 score for the large objects have decreased when using sliced inference.

For the cropped training and standard inference experiments we see that the performance drops compared to the standard training and standard inference experiments.

As expected we see that the large version of the models outperform the smaller version. The smaller models seem to benefit more from sliced inference compared to the larger models.

The YOLOv5-Large model trained on the cropped dataset and evaluated with sliced inference gives us the highest score of 48.8, compared to the corresponding YOLOX-Large model trained on the cropped dataset and evaluated with sliced inference, which gave a AP50 score of 45.2. For all of the experiments, the YOLOv5 models outperformed their corresponding YOLOX models in the same experimental settings.

We see that training the models on the cropped dataset and testing them in the standard inference setting lowers the performance. The models trained on the cropped dataset give a substantial increase in performance when evaluated using sliced inference. The YOLOv5-Small model gave a 4.1 point increase in the AP50 score when sliced inference was used, the YOLOv5-large model trained on the cropped dataset gave an increase of 10.8 points in the AP50 score.

The YoloV5-Large model trained on the cropped dataset and evaluated using sliced inference gives a AP50 score of 54.7 for the pedestrian class. The pedestrian class is the class with the smallest object size in the dataset, and an additional challenge for detecting the pedestrian class in this dataset is that there are many instances of images where there are instances of the pedestrian class occurring in a crowd. While we see that there is increase in performance in all classes in the experiments where sliced inference alone or cropped training and sliced inference were used, the increase in larger objects seem to be behind the increase in the smaller objects. The YOLOv5-Large model trained on the cropped dataset and evaluated using sliced inference resulted in an increase of %12,2 for the pedestrian class while the increase for the truck (or other larger class) class was %6,4, compared to the YOLOv5-Large model which was trained on the standard dataset and evaluated using standard inference.

## VI. CONCLUSIONS

We see that YOLOv5 and YOLOX models give relatively good performance for small object detection problems. We also see that using sliced inference and a special dataset, we can substantially improve the performance of object detection models. From table 1 we see that models trained on the cropped dataset gives a lower AP50 score compared to the corresponding models trained on the standard dataset in standard inference experiments. The YOLOv5 models seem to have higher performance compared to YOLOX models for all small, medium and large object size classes.